\documentclass{article}
\usepackage{graphicx}  

\begin{document}

\title{Big Data Scaling through Metric Mapping: Exploiting the Remarkable Simplicity of 
Very High Dimensional Spaces using Correspondence Analysis}
\author{Fionn Murtagh \\
University of Derby; Goldsmiths University of London \\
Email fmurtagh@acm.org}

\maketitle

\abstract{We present new findings in regard to data analysis in very high dimensional
spaces.  We use dimensionalities up to around one million.  
A particular benefit 
of Correspondence Analysis is its suitability for carrying out an orthonormal 
mapping, or scaling, of power law distributed data.  Power law distributed data are 
found in many domains. Correspondence factor analysis provides a latent semantic or 
principal axes mapping. Our experiments use data from digital chemistry and finance, 
and other statistically generated data.}

\section{Introduction}

Correspondence analysis of an infinite (unbounded) number of rows or observations, 
crossed by 1000 attributes, was discussed in \cite{jpb1982}, and also \cite{jpb1997}.
Our objective in this article is to describe useful properties of data spaces,
of high dimensionality. Our particular interest is
in properties that are of benefit to ``big data'' analytics.  See \cite{murtslds} for
further examples of application.  

It was shown experimentally in 
\cite{mur04} how points in high dimensional spaces become increasingly equidistant
with increase in dimensionality.  Both \cite{donoho} and \cite{hall} study Gaussian
clouds in very high dimensions.  The former finds that ``not only are the points 
[of a Gaussian cloud in very high dimensional space] on the convex hull, but all 
reasonable-sized subsets span faces of the convex hull.  This is
wildly different than the behavior that would be expected by traditional low-dimensional
thinking''.

That very simple structures come about in very high dimensions can have 
far-reaching implications.  Firstly, even very simple structures
(hence with many symmetries) can be used to support fast, and perhaps even constant
time worst case, proximity search \cite{mur04}.  Secondly, as shown in
the machine learning framework by \cite{hall},
there are important implications ensuing from the simple high dimensional structures.
Thirdly, \cite{remarkable} shows that very high dimensional clustered data contain
symmetries that in fact can be exploited to ``read off''
the clusters in a computationally efficient way.  Fourthly, following
\cite{delon}, what we might want to look for in contexts of
considerable symmetry are the ``impurities'' or small irregularities
that detract from the overall dominant picture.

In general, data analysis considered as the search for symmetries in data, is 
discussed in \cite{murstek}.  This relates in particular to hierarchical clustering. 
That can be considered as a natural extension of the work described in this paper. 

\section{Properties of Very High Dimensional Data Spaces}
\label{sect2}

\subsection{Piling and Concentration of Data, with Increase in Dimensionality}

With high dimensional, sparse data \cite{hall}, there is a very strong concentration 
of our clouds (rows/points, columns/projection vectors) into concentrated (i.e.\ small 
variance) Gaussians.   Therefore, there is a good approximation of
our cloud by its mean. This in turn means that the mean random projection is
a very good representative of our data cloud. 

From the given, non-negative valued data, $k_{IJ}$, our $I$ cloud and $J$ cloud 
are converted to frequencies, denoted  $f_{IJ}$ with associated mass distributions,
$f_I$ and $f_J$.  
The conditional distribution of $f_J$ knowing $i \in I$, also termed
the $j$th profile with coordinates indexed by the elements of $I$, is
$$ f^i_J = \{ f^i_j = f_{ij}/f_i = (k_{ij}/k)/(k_i/k) ; f_i \neq 0 ; 
j \in J \}$$ and likewise for $f^j_I$.  Thus our data are points in a 
high dimensional data cloud, defining row or column profiles.

Through high dimensional piling, i.e.\ concentration, we have that the profile
vectors tend towards the average profile.  What gives rise to this is
sparsity through high dimensionality, which also implies low sample (or population)
size.  It implies this because we are not considering here the case of both population
size and dimensionality tending to infinity at the same, or related, rate.

By the central limit theorem, and by the concentration (data piling) effect
of high dimensions \cite{hall,terada}, we have as dimension $m \rightarrow \infty$:
pairwise distances become
equidistant; orientation tends to be uniformly distributed.  We find also: the norms
of the target space axes are Gaussian distributed; and as typifies sparsified data,
the norms of the points in our high dimensional data cloud, in the factor space, 
are distributed as a negative exponential or a power law.

\subsection{Relative and Absolute Contributions}
\index{contribution}

The moment of inertia of the clouds $N_J(I)$ and $N_I(J)$, relative to the 
$\alpha$ axis, is $\lambda_{\alpha}$.   Let  
$\rho$ be the Euclidean distance from the cloud
centre in the factor space, and let the projection of $i \in I$ on the $\alpha$ factor 
be $F_{\alpha}(i)$.  Decomposition of the cloud's inertia is then as follows. 
\begin{equation}
M^2(N_J(I)) = \sum_{\alpha=1..\nu} \lambda_\alpha = \sum_{i \in I} f_i \rho^2(i)
\label{eqn1sect22} 
\end{equation}
In greater detail, we have for this decomposition: 

\begin{equation} 
\lambda_\alpha = \sum_{i \in I} f_i F^2_\alpha(i) \mbox{   and   }
\rho^2(i) = \sum_{\alpha=1..\nu} F^2_\alpha(i) 
\label{eqn2sect22}
\end{equation}

Contributions to inertia are fundamental in order to define the mapping into the
factor space.  Contributions by row points, or by column points, in their respective
dual spaces, define the importance of those given data elements for the constructed
mapping.  Supported by the experimental results to be reported on in the following
sections, we will use the average contribution to the inertia as a measure of 
cloud concentration.  The inertia is the fundamental determinant of not just 
relative positioning, but of essential cloud properties.  

Why we use the contributions to the total inertia of the cloud, as the basis for
a measure of concentration, is motivated for the following reason.
Consider the following hypothetical scenario.  Consider where massive points in the 
cloud were moved towards the centre or origin, leaving light points to drift away from 
the centre.  Through inertia, we would characterize such a scenario as concentration. 
Or consider where massive points drift apart, and their inertia contributions outpace
the inertia contributions of less massive points that move closer to the origin. 
Again in that scenario, our inertia measure of concentration would be appropriate for 
quantifying the lack of concentration.  In these hypothetical scenarios, we see how 
contribution to inertia is a key consideration for us.  Inertia is more important than 
projection (i.e., position) per se.  

We now look at absolute versus relative contributions to inertia.  The former one 
of these is the more relevant for us.  This will be seen in our experiments below. 
What we consider for the attributes (measurements, dimensions) holds analogously 
for the observations.  

\begin{itemize}
\item
$f_j \rho^2(j)$ is the absolute contribution of attribute $j$ to the
inertia of the cloud, $M^2(N_I(J))$, or the variance of point $j$.
Therefore, from expressions \ref{eqn2sect22}, this absolute contribution
of point $j$ is also: $f_j \sum_{\alpha=1..\nu} F^2_\alpha(j)$.  

\item 
$f_j F^2_\alpha(j)$ is the absolute contribution of point $j$ to the
moment of inertia $\lambda_\alpha$.

\item
$f_j F^2_\alpha(j) / \lambda_\alpha $ is the relative contribution of
point $j$ to the moment of inertia $\lambda_\alpha$.  
We noted in \ref{eqn1sect22} that $\lambda_\alpha = \sum_{j \in J} f_j F^2_\alpha(j)$.
So the relative contribution of point $j$ to the moment of inertia $\lambda_\alpha$
is: $f_j F^2_\alpha(j) / \sum_{j \in J} f_j F^2_\alpha(j)$.  The total relative 
contribution 
of $j$, over all $j \in J$, is 1.  The total contribution over all factors, 
indexed by $\alpha$, then becomes $\nu$, the number of factors.  So the mean 
contribution (here, the mean relative contribution) of the attributes, is 
$\frac{\nu}{|J|}$.  In the simulations below, 
the trivial first eigenvalue, and associated axis, is included here.  
\end{itemize}

We have now the technical machinery needed to evaluate data clouds in very 
high dimensions.  We will keep our cloud of observables, small.  This is 
$N(I)$.  It is in a $|J|$-dimensional space.  That dimensionality, $|J|$, 
will be very large.  That is to say, the cloud of what we take as attributes, $N(J)$,
will be huge.  While the cloud itself, $N(J)$, is huge, each point in that cloud, 
$j \in J$ is in a space of dimension $|I|$, which is not large.  

Now we will carry out our evaluations.  Our choice of cloud cardinality and 
dimensionality are motivated by inter-study comparison.  
The R code used is available at the web site, {\tt www.correspondances.info}.

\section{Evaluation 1: 
Uniformly Distributed Points in Data Clouds of Dimensionality up to One 
Million}
\label{eval1}

Uniformly distributed values, in $[0,1]$, were used for five data clouds, each 
of 86 points in dimensionalities of: 100, 1,000, 10,000, 100,000 and 1,000,000.  In 
the usual analysis perspective, we have 86 observations, and the dimensionalities 
are associated with the attributes or features.  This 
input data is therefore dense in value.  Results 
obtained are shown in Table \ref{tab1}.  

Note how increasing dimensionality implies 
the following. We base concentration, or compactness, on the absolute contribution 
to the inertia of the factors.
The average absolute contribution to the factors tends towards zero.  The standard 
deviation also approaches zero. Thus the cloud becomes more compact.  

We provide 
median as well as mean as an indication of distributional characteristics of the
absolute contribution that we are examining.  We observe a relatively close match between
mean and median values, implying an approximate Gaussian distribution of the 
absolute contributions.  
For all cases (including the 1,000,000-dimensional case), we checked that the 
distributions of absolute and relative contributions, and norms squared of the input data,
are, visually, close to Gaussian.

The maximum projection values, that do not decrease, serve to show that concentration 
with increasing dimensionality is a phenomenon relating to the whole cloud, and therefore 
to the average (or median).   

\begin{table}
\caption{Five data clouds, each of 86 points in spaces of dimensionality:
100, 1,000, 10,000, 100,000 and 1,000,000. The original coordinate values are randomly 
uniform in $[0,1]$.}
\begin{center}
\begin{tabular}{|rclll|} \hline
\multicolumn{5}{c}{Contributions to Inertia of Factors by the Columns} \\ \hline
Dim.  &  Contributions &  Mean        &  Std.Dev.   & Median  \\ \hline 
100  &  Absolute      &  0.01322144   & 0.0005623589 & 0.01325343 \\
     &  Relative      &  0.86         & 0.04588791   & 0.869127 \\
1000 &  Absolute      &  0.001331763  & 5.440168e-05 & 0.001333466 \\
     &  Relative      &  0.086        & 0.009729907  & 0.08547353 \\
10000 & Absolute      &  0.0001332053 & 5.279421e-06 & 0.0001332981  \\
      & Relative      &  0.0086       & 0.0009742588 & 0.008577748 \\
100000 & Absolute     &  1.330499e-05 & 5.269165e-07 & 1.332146e-05 \\
       & Relative     &   0.00086     & 9.783086e-05 & 0.0008574684 \\
1000000 & Absolute    &  1.330706e-06 & 5.278487e-08 & 1.332186e-06 \\
        & Relative    &  8.6e-05      & 9.788593e-06 & 8.576992e-05 \\ \hline
\end{tabular}
\vskip 0.5cm 
\begin{tabular}{|rl|} \hline
\multicolumn{2}{c}{Maximum factor projection} \\
Dim.  & Projection \\ \hline
 100  &  0.3590788 \\
1000  &  0.2777193 \\
10000 &  0.2799913 \\
100000 & 0.3678137 \\
1000000 &  0.3750852 \\ \hline
\end{tabular}
\end{center}
\label{tab1}
\end{table}

\subsection{Computational Time Requirements}

The largest, uniformly random generated, dataset used was of dimensions 
$86 \times 1000000$.  In order to create this data array, 
an elapsed time of 82.8 seconds was required.  Carrying out the main processing, 
furnishing the results in Table \ref{tab1}, involved a basic Correspondence
Analysis of this input data matrix. The projections and contributions (to inertia)
of the 86 points were to be determined.

Standard processing proved satisfactory for
these evaluations.  For this large data set, our main processing took an 
elapsed time of 95.6 seconds. 

Our machine used was a MacBook Air, with a 2 GHz processor, and 8 GB of memory,
running OS X version 10.9.4.  The version of R in use was 2.15.2.

\section{Evaluation 2: Time Series of Financial Futures in Varying Embedding 
Dimensions}
\label{eval2}

The following data were used in \cite{remarkable}.  In that work 
we used the sliding window approach to embed the financial signal in 
spaces of varying dimensionality.  The work in \cite{remarkable} showed, 
in various examples, how there may be no ``curse of dimensionality'', in Belman's
\cite{bellman} famous phrase, in very high dimensional spaces.  There is no
such obstacle if we seek out, and make use of, the ``remarkable simplicity'' 
\cite{remarkable} of very high dimensional data clouds.  

We use financial futures, from circa March 2007, denominated in euros
from the DAX exchange. Our data stream, at the millisecond rate, comprised 
382,860 records. Each record includes: 5 bid and 5 asking
prices, together with bid and asking sizes in all cases, and action. 

We extracted one symbol (commodity) with 95,011 single bid values, on which
we now report results.  These values were continuous and avoided missing values.
The data values were between 6788 and 6859.5 in value.  
There were either integer valued, or ending in 0.5.  Very often this signal contained
short sequences of successive identical values.  

Similar to \cite{remarkable}, we define embeddings of this financial signal as follows. 
Each embedding begins at the following time steps in the financial signal: 1, 1000, 
2000, $\dots$, 85000.  The lengths of the success embeddings were, in our three
case studies: 100, 1000, 10000.  That provided matrices, in these three case studies,
of sizes: $86 \times 100, 86 \times 1000, 86 \times 10000$. 

Results obtained are presented in Table \ref{tab2}.  The histograms of projections
on the factors were visually observed to be Gaussian-distributed.  We observe
how the mean absolute contribution, as well as the median absolute contribution,
decrease as the embedding dimensionality increases.  The standard deviation of
absolute and of relative contributions decrease too, indicating the increasing
concentration.  Our measure of concentration is the average (or median) contribution
by the embedding dimensionality values (what we may consider as attributes or
characterizing features of the ``sliding window'' over the signal) to the inertia of
the factors.  We observe also how the maximum projection on the factors does not
decrease.  This just means that the cloud in the overall sense, and on the whole,
gets increasingly compact or concentrated, as the attribute dimensionality increases.

\begin{table}
\caption{Embeddings, of dimensionalities 100, 1000 and 10,000, for a financial 
time series.}
\begin{center}
\begin{tabular}{|rllll|} \hline 
Dim.  &  Contribution &  Mean  & Std.Dev.  &  Median \\ \hline
100   & Absolute   &    0.01  & 9.260615e-08 &  0.01000002 \\
      & Relative   &    0.86  & 0.05399462   &  0.8672608 \\
1000  & Absolute   &    0.001 & 3.297399e-08 &  0.001000008 \\
      & Relative   &    0.086 & 0.0121773   &  0.08518253 \\
10000 & Absolute   &    0.0001000001 & 2.168381e-08 & 9.999872e-05 \\
      & Relative   &    0.0086 & 0.001159708  &  0.008477465 \\ \hline
\end{tabular}
\vskip 0.5cm
\begin{tabular}{|rl|} \hline
\multicolumn{2}{|c|}{Maximum factor projection} \\
  Dim.  & Projection \\  \hline
 100 &  0.0001054615 \\
1000 & 0.0002979516 \\ 
10000 & 0.0008869227 \\ \hline
\end{tabular}
\end{center}
\label{tab2}
\end{table}

\section{Evaluation 3: Chemistry Data, Description of Its Power Law Property}
\label{eval3}

\subsection{Data and Determining Power Law Properties}

The following data were used in our earlier work in \cite{chemical}.
We used a set of 1,219,553 chemical structures coded through 1052
presence/absence values, using the Digital Chemistry bci1052 dictionary
of fragments \cite{wright}. That binary-valued matrix was sparse: occupancy 
(i.e.\ presence = 1 values) 
of the chemicals crossed by attribute values was 8.6\%.  

Our motivation here is to investigate the effect of greatly increasing the 
attribute dimension.  In the next section we will develop a novel way to do this. In 
this section we determine the relevant statistical properties of our data. 

Here, we will use 425 chemicals from this set, 
in 1052-dimensional space.  We took 425 chemicals in order to have a limited set,
$|I| = 425$, in the attribute space, $J$.  Each chemical had therefore presence/absence
(i.e.\ 1 or 0, respectively) values on $|J| = 1052$ attributes.  The occupancy of 
the 425 $\times$ 1052 data set used was 5.9\%.  Since we wanted this sample of 
425 of the chemicals to be representative of the larger set from which they came, 
we now look at the most important distributional properties.

\begin{figure}
\centering
\includegraphics[width=6cm]{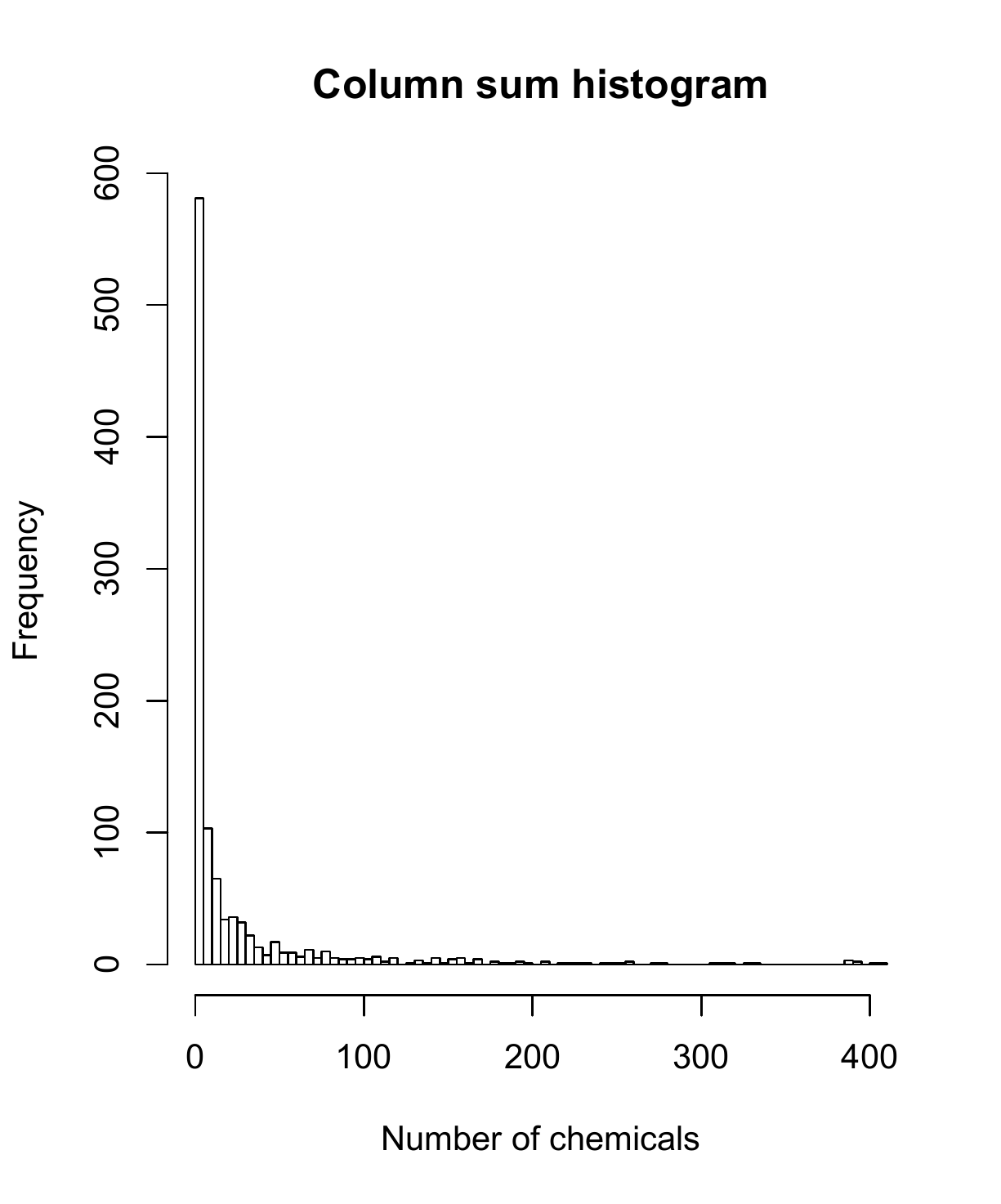}
\caption{Histogram of column, i.e.\ chemical attribute, sums.}
\label{fig1}
\end{figure}

The marginal distribution, 
shown in Figure \ref{fig1}, is not unlike the marginal distribution displayed
in \cite{chemical}.  In that previous work, we found the power law distribution 
of the chemical attributes to be of exponent $-1.23$.  Let us look at the 
power law of the baseline distribution function used here, i.e.\ relating to the 
425 chemicals. 

A power law (see \cite{mitz}) is a frequency of occurrence distribution of 
 the general form $x^{-\alpha}$ where constant $\alpha > 0$; whereas an exponential law is
of the form $e^{-x}$.  For a power law, the probability that a value, following 
the distribution, is greater 
than a fixed value is as follows: $P(x > x_0) \sim c x^{-\alpha}$,
$c, \alpha > 0$.  A power law has heavier tails than an exponential
distribution.  In practice, $0 \leq \alpha \leq 3$.  For such values,
$x$ has infinite (i.e.\ arbitrarily large)
variance; and if $\alpha \leq 1$ then the mean of $x$ is
infinite.  The density function of a power law is $f(x) = \alpha c
x^{-\alpha - 1}$, and so $\ln f(x) = - \alpha \ln x  + C$, where $C$ is a
constant offset.  Hence a log-log plot shows a power law as linear.  Power
laws have been of great importance for modelling networks and other
complex data sets.  

\begin{figure}  
\centering
\includegraphics[width=6cm]{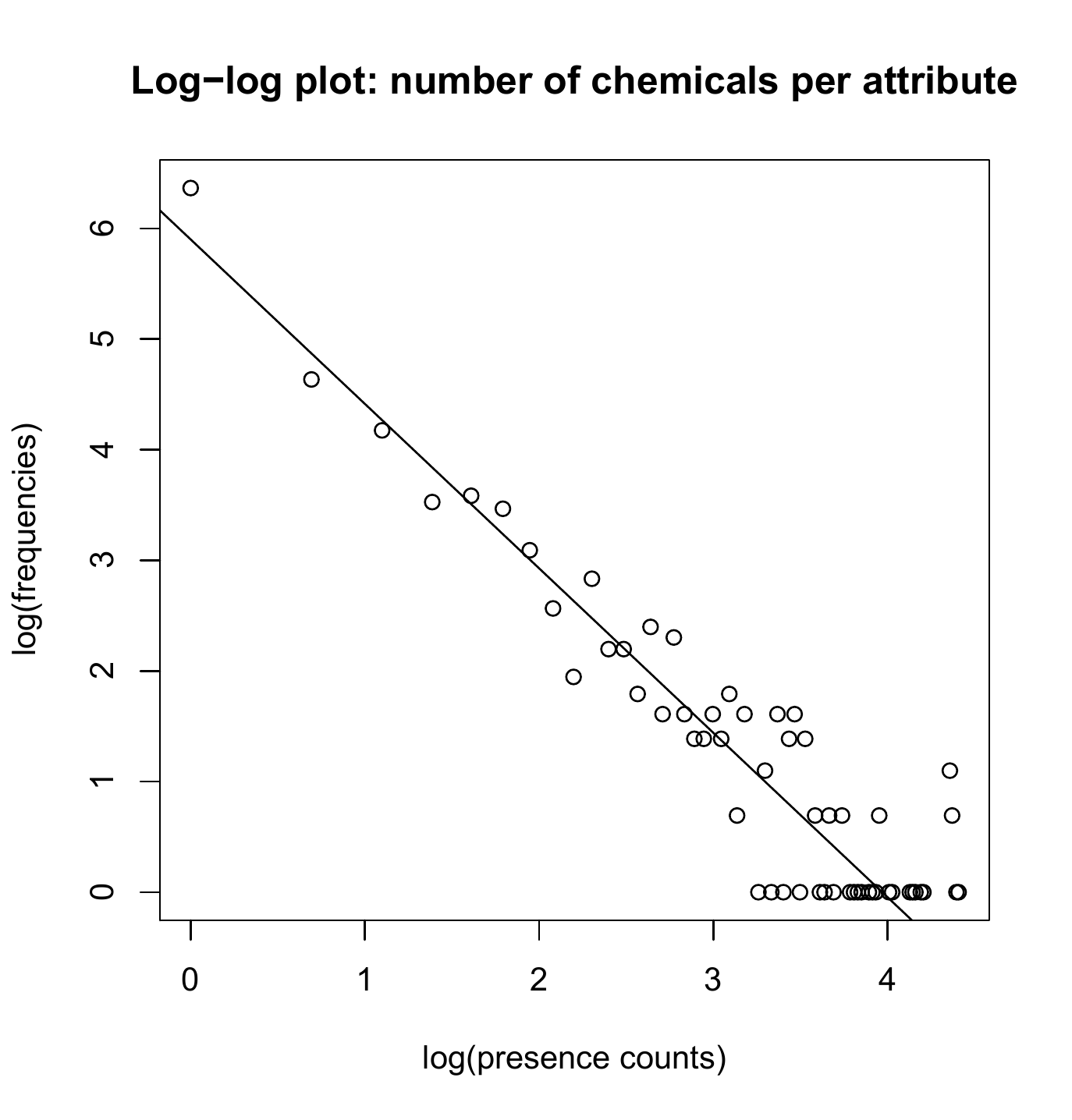}
\caption{Log-log plot of numbers of chemicals per attribute, based on the data set of 
425 chemicals.}
\label{fig2}
\end{figure}

Figure \ref{fig2} shows a log-log plot based on the 1052 presence/absence attributes,
using the 425 chemicals.   In a very similar way to the power law properties
of large networks (or file sizes, etc.) we find an approximately linear
regime, ending (at the lower right) in a large fan-out region.
The slope of the linear region characterizes the power law.  For this data,
we find that the probability of having more than $n$ chemicals per attribute to be
approximately $c/n^{1.49}$ for large $n$.

The histogram of attributes per chemical, on the other hand, is approximately a 
Gaussian. This is as observed in \cite{chemical}.

\subsection{Randomly Generating Power Law Distributed Data in Varying Embedding Dimensions}

In section \ref{eval1} we used dense uniformly distributed data.  In section \ref{eval2},
our financial futures were slow-moving, in the sense of small variation between successive
values.  But there too the data were dense and real-valued.  Our chemistry context is sparse
and boolean-valued (for presence/absence).  We use this context to generate data that 
keep the property of the attributes (i.e., the 
columns or dimensions) following a power law in regard to their distribution.  

To generate new random data sets that fully respect the distributional characteristics
of our known data, we will use the distribution function that is displayed in 
Figure \ref{fig1}.  This is the data distribution of coding attributes that characterize
the chemicals, i.e.\ presence of molecules.  

In line with our earlier notation, the marginal distribution in Figure \ref{fig1} 
is $f_J$ for attribute set, $J$.  The chemicals set is $I$.  The presence/absence 
cross-tabulation of chemicals by their attributes is, in frequency terms, $f_{IJ}$.   
The $(i,j)$ elements, again in frequency terms, is $f_{i,j}$.  In whole number terms,
representing presence or absence, i.e.\ 1 or 0, the chemicals-attributes cross-tabulation
is denoted $k_{IJ}$.  

We generate a new data set that cross-tabulates a generated set of
chemicals, $I'$, crossed by a generated set of attributes, $J'$.  Let $|.|$ denote
cardinality.  We randomly sample (uniformly) $|J'|$ values from $k_J$.  Therefore 
we are constructing a new, generated set of attribute marginal sums.  The generated
values are of the same distribution function. That is,
both $f_{J'} \sim f_J$ and $k_{J'} \sim k_J$.  
The next step is to consider the newly generated chemicals, in the set $I'$, of 
cardinality $|I'|$.  Given $k_{j'}$, we generate $|k_{j'}|$ values of 1 in the set
of $|I'|$ elements.  In this way, we generate the chemicals that contribute the 
$k_{j'}$ attribute presences found for attribute $j'$.   

For the generated chemical data, we use 425 chemicals, in attribute spaces of 
dimensions 1052, and then, 10 times this, 100 times this, and 1000 times this 
dimensionality.  

See the R code used at {\tt www.correspondances.info} 
(see under ``Evaluation 3'').  This code shows 
the case of 1000 times the dimensionality.  I.e., for 425 chemicals with 1052 
presence/absence or one/zero values, we generate a matrix of 
425 chemicals $\times$ 1052,000 presence/absence attributes.  For the $425 \times 
1052$ matrix, we have 26405 presence values, and 
a density (i.e., presence or 1 values) of 5.9\%.  For the 
generated $425 \times 1052000$ presence/absence attributes, we have 5645075 presence
values, and a density of 1.26\%.   

\begin{figure}
\centering
\includegraphics[width=12cm]{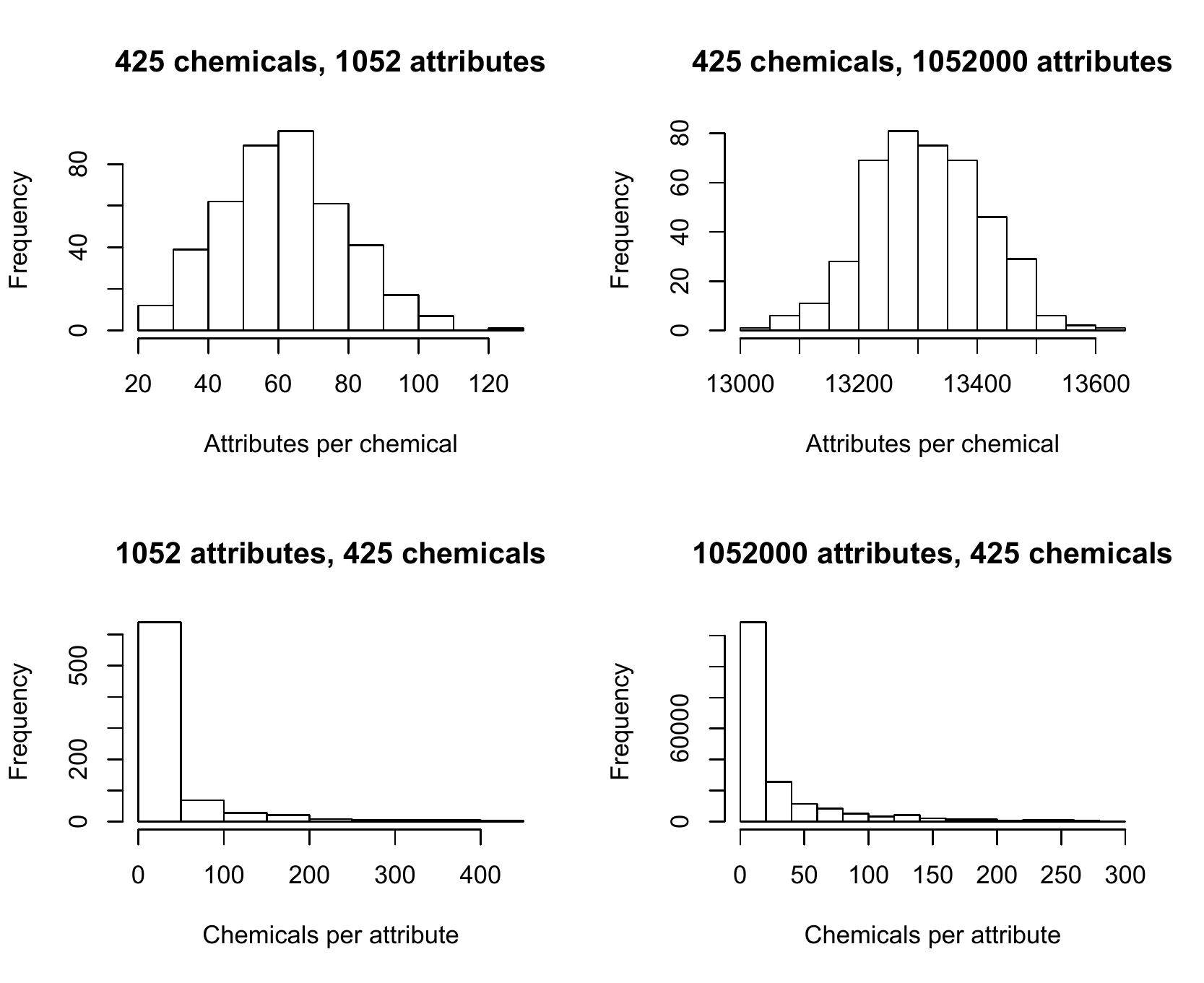}
\caption{Histograms of marginal distributions of the original $425 \times 
1052$ chemicals by attributes, and the generated data with similar marginal distributions,
of $425 \times 1052000$ chemicals by attributes.  Marginal distribution values greater
than 0 were taken into account.}
\label{fig3}
\end{figure}

Figure \ref{fig3} displays the marginal distributions.  This shows visually how
well our generated data approximates the original data.  Let us also look at how 
close the power law distributional properties are.  
Table \ref{tab3} lists the power law exponents for our generated data sets.  

\begin{table}
\caption{Power law exponents for generated chemical data, with 425 chemicals, 
with presence/absence (respectively 1 or 0) in attribute dimensions: 
1,052, 10,520, 105,200 and 1,025,000.}
\begin{center}
\begin{tabular}{|rl|} \hline
\multicolumn{2}{|c|}{425 chemicals} \\ \hline
Dim.      &   Exponent \\ \hline
1052      &  -1.49 \\
10520     &  -1.75 \\ 
105200    &  -1.64 \\
1052000   &  -1.78 \\ \hline
\end{tabular}
\end{center}
\label{tab3}
\end{table}

\begin{table}
\caption{425 chemicals with presence/absence values on the following 
numbers of characterizing attributes: 1,052, 10,520, 105,200 and 1,052,000.
The dimensionality of the space in which the chemicals are located is 
given by the number of characterizing attributes.}
\begin{center}
\begin{tabular}{|r|ll|l|} \hline 
425 chemicals    & \multicolumn{2}{c}{Absolute contribution} &   \\
Dimensionality &  Mean & Std.Dev. & Max. projection \\ \hline  
1052     &  0.01161321 & 0.007522956 & 16.27975 \\
10520    &  0.00133034 & 0.002798697 & 12.31945 \\
105200   &  0.000140571 & 0.0002946923 & 10.91465 \\
1052000 &  1.39319e-05 & 2.919471e-05 & 11.06306 \\ \hline
\end{tabular}
\end{center}
\label{tab4}
\end{table}

Table \ref{tab4} shows clearly how the absolute contribution to the inertia of 
the factors, which is mass times distance squared, becomes of smaller mean value,
and of smaller standard deviation (hence the mean is a tighter estimate), as 
dimensionality increases.  The degree of 
decrease of the mean value is approximately linear in the increase of dimensionality
(i.e.\ tenfold for each row of Table \ref{tab4}).  Once again, we show very conclusively
how increasing dimensionality brings about a very pronounced concentration of the 
data cloud that is considered.  As dimensionality increases, the cloud becomes 
much more compact, i.e.\ far more concentrated.

\section{Conclusion}
\label{concl}

We explored a wide range of evaluation settings. 
We have shown that is is easy and straightforward to analyze 
data that are in very high attribute dimensions (or feature dimensions, in other 
words, typically the number of columns of our input data matrix).  Of course one
needs to understand the nature of one's analysis.  It is not a ``black box'' process.  
Instead it is necessary to investigate how to ``let the data speak''.

\end{document}